\documentclass[10pt,twocolumn,letterpaper]{article}

\usepackage[pagenumbers]{cvpr} 

\usepackage{algorithm}

\definecolor{cvprblue}{rgb}{0.21,0.49,0.74}
\usepackage[pagebackref,breaklinks,colorlinks,allcolors=cvprblue]{hyperref}
\def\paperID{1996} 
\def\confName{CVPR}
\def\confYear{2025}
\title{HiFiVFS: High Fidelity Video Face Swapping }

\author{
\textbf{Xu Chen}\textsuperscript{1*}
\ 
\textbf{Keke He}\textsuperscript{1*}
\ 
\textbf{Junwei Zhu}\textsuperscript{1†}
\ 
\textbf{Yanhao Ge}\textsuperscript{2}
\ 
\textbf{Wei Li}\textsuperscript{2} 
\ 
\textbf{Chengjie Wang}\textsuperscript{1}
\\
\textsuperscript{1} Tencent
\quad
\textsuperscript{2} VIVO \\
\url{https://cxcx1996.github.io/HiFiVFS/}
}

\begin{document}
\twocolumn[{ 
\renewcommand\twocolumn[1][]{#1}
\maketitle 
\begin{figure}[H] 
\hsize=\textwidth
\centering
\vspace{-2em}
\includegraphics[width=0.99\textwidth]{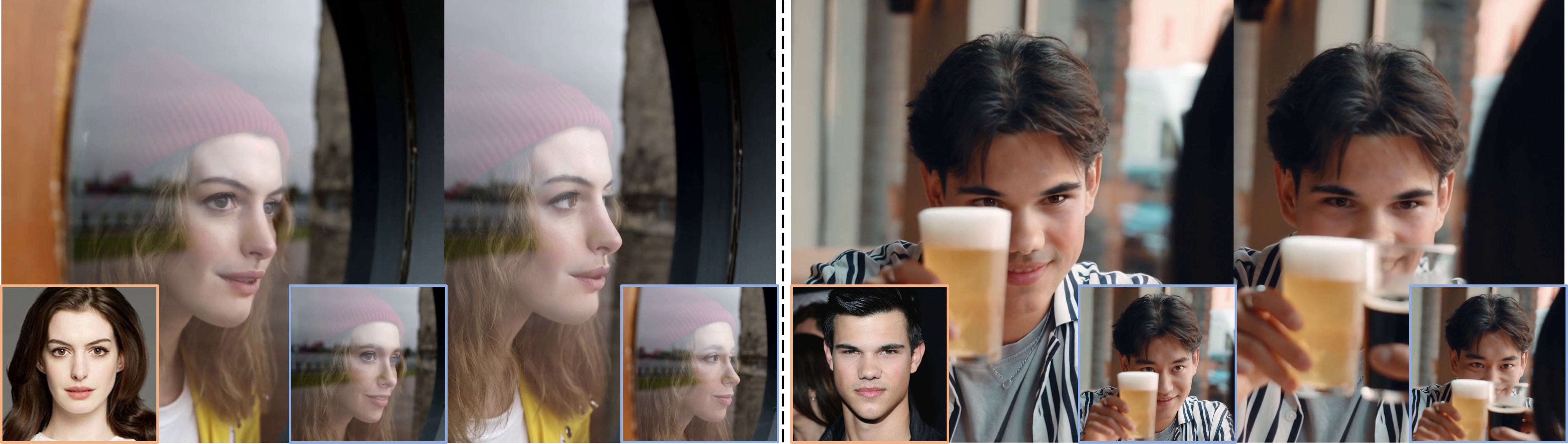}
\caption{
\textbf{Face swapping results of HiFiVFS.} The face in the source image (orange) is taken to replace the face in the target video (blue).}
\label{fig:fig1}
\end{figure} 
}]

\footnotetext[1]{Equal Contribution.}  
\footnotetext[2]{Corresponding Author.}
\begin{abstract}
Face swapping aims to generate results that combine the identity from the source with attributes from the target. Existing methods primarily focus on image-based face swapping. When processing videos, each frame is handled independently, making it difficult to ensure temporal stability. From a model perspective, face swapping is gradually shifting from generative adversarial networks (GANs) to diffusion models (DMs), as DMs have been shown to possess stronger generative capabilities. Current diffusion-based approaches often employ inpainting techniques, which struggle to preserve fine-grained attributes like lighting and makeup. To address these challenges, we propose a high fidelity video face swapping (HiFiVFS) framework, which leverages the strong generative capability and temporal prior of Stable Video Diffusion (SVD). We build a fine-grained attribute module to extract identity-disentangled and fine-grained attribute features through identity desensitization and adversarial learning. Additionally, We introduce detailed identity injection to further enhance identity similarity. Extensive experiments demonstrate that our method achieves state-of-the-art (SOTA) in video face swapping, both qualitatively and quantitatively.

\end{abstract}    
\section{Introduction}
\label{sec:intro}

\begin{figure*}[t]
  \centering
   \includegraphics[width=0.99\linewidth]{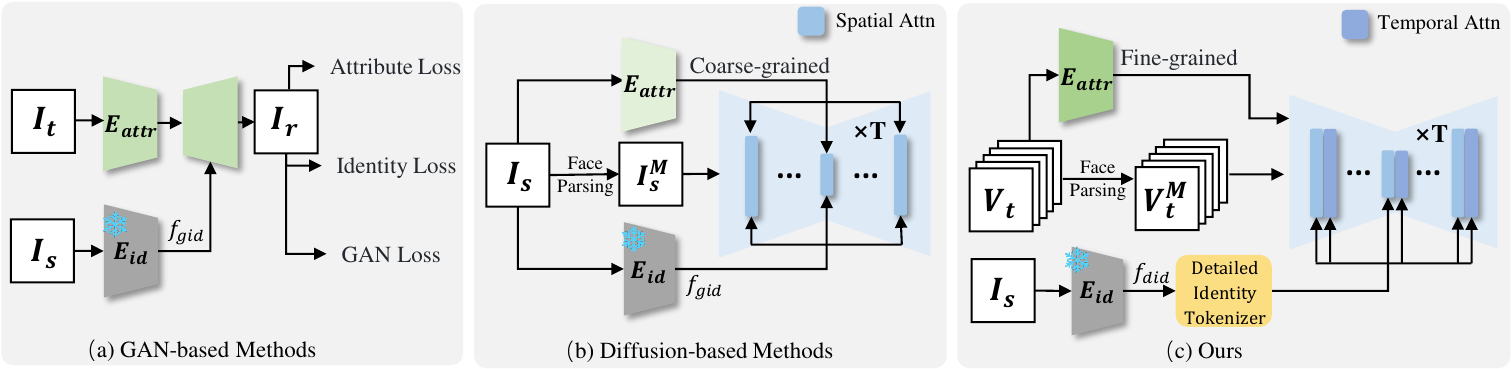}
 
   \caption{Training pipeline of face swapping methods. (a) GAN-based methods achieve feature disentanglement by using attribute and identity loss along with adversarial learning. (b) Diffusion-based methods construct an inpainting data flow that leverages pre-trained identity and attribute features to fill in facial areas. (c) Our HiFiVFS is designed for video face swapping by incorporating temporal attention on multiple frames and introducing temporal identity injection. We also introduce a fine-grained attribute extractor and a detailed identity tokenizer to improve control over attributes and identities.
   }
   \label{fig:ppl_cmp}
\end{figure*}

Face swapping involves generating an image or video that combines the identity of a source face with the attributes (such as pose, expression, lighting, and background) from a target image or video, as illustrated in Fig.~\ref{fig:fig1}. This technique has attracted significant interest due to its potential applications in the film industry~\cite{alexander2009creating}, video games, and privacy protection~\cite{ross2010visual}. 
Existing studies on face swapping have primarily concentrated on still images. To process a video, each frame is handled as an individual image before being compiled into the final video.
This approach overlooks the continuity across multiple frames, potentially leading to temporal jitter in the output. Moreover, current methods have not sufficiently tackled the challenges of managing fine-grained attributes, identity control, and maintaining high-quality generation simultaneously.

Existing face swapping methods can be simply categorized into two main types: GAN-based~\cite{chen2020simswap,wang2021hififace,li2019faceshifter,gao2021information,shiohara2023blendface,xu2022MobileFaceSwap,Xu_2022_CVPR,xu2022designing,liu2023fine,kim2022smooth,shu2022few,li2021faceinpainter,xu2022high} and diffusion-based~\cite{zhao2023diffswap,liu2024towards,han2024face} approaches. As shown in Fig.~\ref{fig:ppl_cmp}(a) GAN-based methods perform face swapping by extracting the global identity feature $f_{gid}$ from the source image $I_s$ and injecting them into generative models. 
GAN-based methods do not require paired ground truth during training and enable the direct application of constraints on attributes and identity, thereby offering enhanced control over these elements.
However, previous GAN-based face swapping techniques have limited generative capabilities, resulting in unsatisfactory outputs in challenging scenarios, such as extreme poses or significant facial structure differences.  

Diffusion-based methods take advantage of the strong generative abilities of diffusion models, producing high-quality outputs.  
However, diffusion-based methods depend on ground truth to generate noised inputs during training. Furthermore, acquiring cross-identity ground truth (images of various individuals sharing the same attributes) is often deemed impractical in face swapping tasks. 
As a result, current diffusion-based methods primarily employ inpainting techniques for learning. As shown in Fig.~\ref{fig:ppl_cmp}(b), during training, both attribute and identity features are extracted from source image $I_s$ and injected into the diffusion model to reconstruct the complete facial image. During inference, attribute feature $f_{attr}$ is extracted from the target image $I_t$, while identity feature $f_{gid}$ is drawn from the source image $I_s$. To prevent the leakage of identity information into attribute features during training, $3$D landmarks~\cite{deng2019accurate, zhao2023diffswap} or CLIP features ~\cite{radford2021clip, han2024face} are commonly used in attribute injection. However, these methods are not suitable for fine-grained attribute control, such as lighting and makeup.

In this work, we proposed a video face swapping framework via diffusion models in Fig.~\ref{fig:ppl_cmp}(c), called HiFiVFS. 
We adopt the Stable Video Diffusion(SVD)~\cite{blattmann2023svd} to address the video face swapping task by incorporating temporal attention on multiple target frames and introducing temporal identity injection.
As a diffusion-based method, the primary challenge lies in obtaining detailed attributes. Unlike previous approaches that directly utilize pre-trained attribute models, HiFiVFS introduces Fine-grained Attributes Learning (FAL), incorporating identity desensitization and adversarial learning into the attribute feature extraction process. This enables our system to acquire detailed attribute features that are independent of identity. Additionally, to reduce the gap between face recognition and face swapping, we designed to introduce Detailed Identity Learning (DIL), which employs detailed identity features that are better suited for the face swapping task and contain richer information, thereby improving the similarity of the face swapping results. We conducted extensive experiments on FaceForensics++~\cite{rossler2019faceforensics++} and 
VFHQ-FS, which include a variety of challenging scenarios such as extreme poses, facial expressions, lighting conditions, makeup, and occlusion, in order to evaluate the effectiveness of our model. Results on various cases show that our method can generate high fidelity and stable results that can better preserve the identity of the source face and detail attribute information of target videos.  Our contributions can be summarized as follows:
\begin{enumerate}
\item  We introduce a high fidelity video face swapping method called HiFiVFS, which can consistently generate high fidelity face swapping videos even in extremely challenging scenarios. To the best of our knowledge, this is the first attempt to improve temporal stability within the face swapping framework.
\item  We proposed Fine-grained Attributes Learning (FAL) and Detailed Identity Learning (DIL) on top of our main model, greatly enhancing the detailed attribute and identity control capability. 
\item  Extensive experiments demonstrate that HiFiVFS outperforms other SOTA face swapping methods on wild face videos across various scenarios.
\end{enumerate}

\section{Related Work}
\label{sec:related}

\subsection{GAN-based methods}

The achievements of generative adversarial networks (GANs)~\cite{goodfellow2014generative,isola2017image} in computer vision have encouraged a lot of research into the face swapping task. Most of these methods~\cite{Nirkin2019fsgan,chen2020simswap,li2019faceshifter,gao2021information,wang2021hififace,zhu2021one,xu2022styleswap,shiohara2023blendface,luo2022styleface,rosberg2023facedancer} adopt a target-oriented pipeline, where an encoder is trained to extract attribute features from the target image. While during the decoding process, identity features extracted from the source image by a pre-trained face recognition model are gradually integrated. SimSwap~\cite{chen2020simswap} introduces a weak feature matching loss in the discriminator's feature space, balancing the preservation of the source identity and the target attributes. FaceShifter~\cite{li2019faceshifter} presents a two-stage framework, where the second stage is designed to correct occlusion artifacts generated by the initial face swap stage. InfoSwap~\cite{gao2021information} employs the Information Bottleneck principle and utilizes information-theoretical loss functions to effectively disentangle identities. 
HifiFace~\cite{wang2021hififace} incorporates a $3$D Morphable Model ($3$DMM)~\cite{blanz1999morphable, thies2016face2face} and integrates a pre-trained $3$D face reconstruction model~\cite{deng2019accurate} into its identity extraction process, which aids in refining the source face shape. BlendFace~\cite{shiohara2023blendface} addresses the attribute biases present in face recognition models and offers well-disentangled identity features specifically designed for face swapping.

However, due to the limited capacity of most GAN models, their performance is not always satisfactory, particularly in challenging scenarios that involve extreme poses and significant variations in facial shape. Additionally, their approach relies on single-frame images for inference, which lacks information from adjacent frames, leading to compromised video stability. In contrast, our method is diffusion-based, providing enhanced generative capabilities to address a broader range of difficult scenarios. Moreover, it operates within a multi-frame framework, resulting in improved temporal stability.

\subsection{Diffusion-based methods}

Diffusion-based methods~\cite{dhariwal2021diffusion,rombach2022high,ye2023ip,xiao2024fastcomposer,wang2024instantid,peng2024portraitbooth} utilize the generative capabilities of the diffusion model to enhance sample quality. To avoid the issue of ID leakage, current diffusion-based methods~\cite{zhao2023diffswap,liu2024towards, han2024face} are all based on an inpainting framework. The target image, with the facial area masked out, is combined with noise and fed into the U-Net. During the denoising process, the UNet gradually incorporates identity features and global attributes such as pose and expression.

Diffswap~\cite{zhao2023diffswap} refers to HifiFace~\cite{wang2021hififace} by introducing $3$DMM~\cite{blanz1999morphable,thies2016face2face} to generate a $3$D keypoint map, allowing the swapped face to maintain the expression and pose of the target. DiffSfSR~\cite{liu2024towards} decomposes the face swapping target into three components: background, identity, and expression. It ensures the consistency of the expression in the swapped result by incorporating a fine-grained expression representation network. However, both methods often exhibit significant differences in detail attributes such as lighting and makeup compared to the template image due to the lack of fine-grained attribute information from the template.

FaceAdapter~\cite{han2024face} not only incorporates $3$D keypoints but also utilizes an additional fine-tuning CLIP to extract features from the target, resulting in better attribute preservation. However, the CLIP model compresses the attribute features significantly, making it unable to express details such as makeup and tattoos on the face. Additionally, it may lead to some leakage of identity features, causing deviations in ID similarity. To address this, our HiFiVFS incorporates disentangled learning into the attribute extraction process, which not only provides complete template attribute features but also prevents the leakage of template ID features.

Additionally, all existing diffusion-based face swapping methods are image-based and fail to achieve temporal stability in video face swapping. In contrast, our approach is specifically designed for video face swapping, featuring a pipeline that supports multi-frame input and output, along with a temporal module to ensure stability across frames.


\section{Preliminary: Stable Video Diffusion}

Stable Video Diffusion (SVD)~\cite{blattmann2023svd} is an advanced latent video diffusion model that excels in high-resolution text-to-video and image-to-video synthesis. To ensure video stability, SVD employs 3D convolution layers, temporal attention layers, and temporal decoder, as presented in~\cite{blattmann2023align}.  

As for training, SVD follow EDM~\cite{karras2022elucidating} framework and precondition the neural network with a dependent skip connection, parameterizing the learnable denoiser $D_{\theta}$ as:
{
\begin{equation} 
D_{\theta}(x; \sigma) = c_{\text{skip}}(\sigma) \, x + c_{\text{o}}(\sigma) \, F_{\theta}(c_{\text{i}}(\sigma) \, x; \, c_{\text{noise}}(\sigma)),
\end{equation}
}
where \( F_{\theta} \) is the model to be trained, \( c_{\text{skip}}(\sigma) \) modulates the skip connection, \( c_{\text{i}}(\sigma) \) and \( c_{\text{o}}(\sigma) \) scale the input and output magnitudes, and \( c_{\text{noise}}(\sigma) \) maps noise level \(\sigma\) into a conditioning input for \( F_{\theta} \). The denoiser $D_{\theta}$  can be trained via denoising score matching (DSM)
{
\begin{equation}
\begin{aligned}
    L_{DM} = \mathbb{E}_{(x_0, c) \sim p_{\text{data}}(x_0, c), (\sigma, n) \sim p(\sigma, n)} \\ \left[ \lambda_\sigma \| D_{\theta}(x_0 + n; \sigma, c) - x_0 \|_2^2 \right],
\end{aligned}
\end{equation}\label{eq:dm}
}
and  \( p(\sigma, n) = p(\sigma) \mathcal{N}(n; 0, \sigma^2) \), \( p(\sigma) \) is a distribution over noise levels \(\sigma\), \(\lambda_\sigma: \mathbb{R}_+ \rightarrow \mathbb{R}_+ \) is a weighting function, and \( c \) is a conditioning signal, such as a text prompt.

\begin{figure*}[t]
  \centering
   \includegraphics[width=0.90\linewidth]{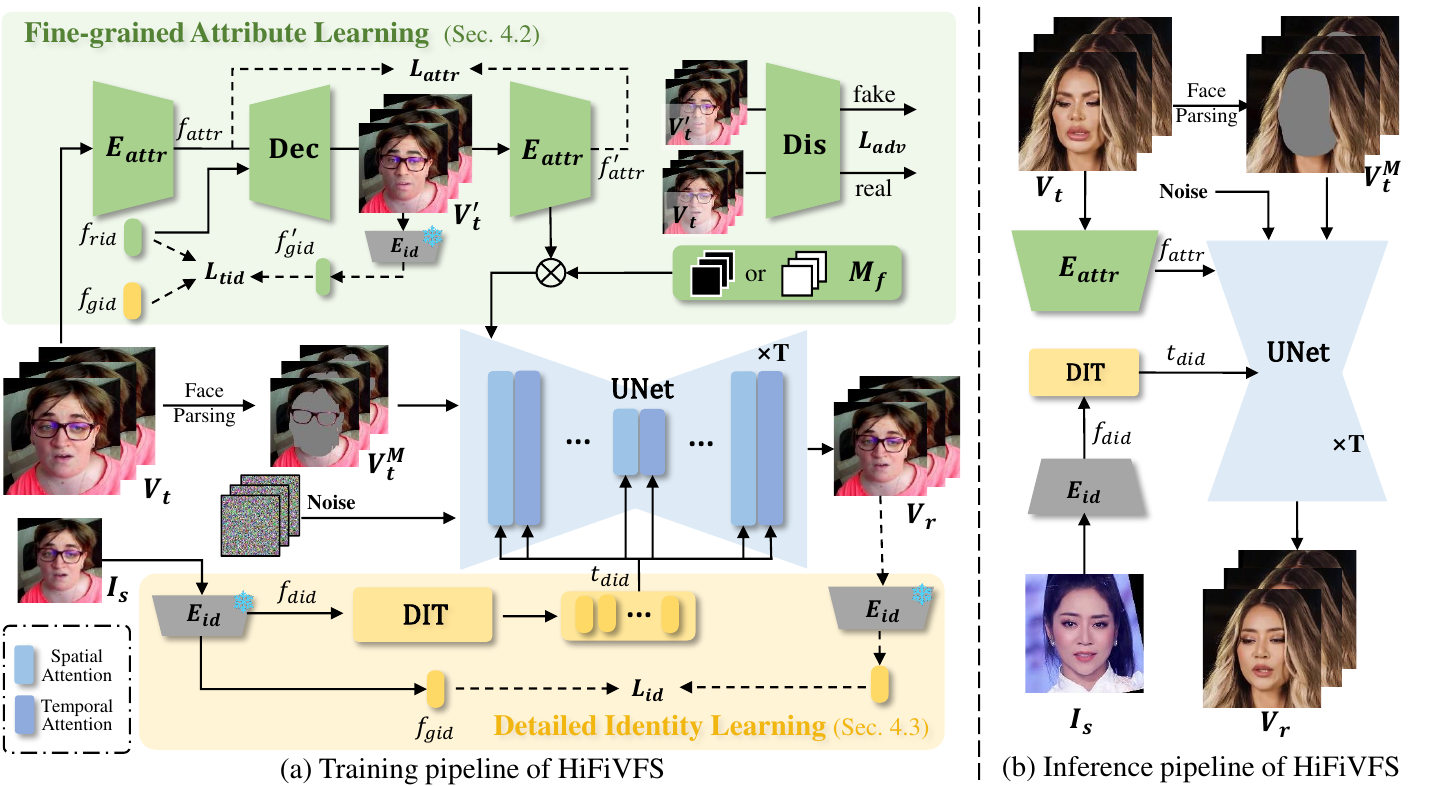}
   \caption{\textbf{Pipeline of our proposed HiFiVFS}, including training and inference phases.
   HiFiVFS is primarily trained based on the SVD~\cite{blattmann2023svd} framework, utilizing multi-frame input and a temporal attention to ensure the stability of the generated videos. 
   In the training phase,  HiFiVFS introduces fine-grained attribute learning (FAL) and detailed identity learning (DIL). In FAL, attribute disentanglement and enhancement are achieved through identity desensitization and adversarial learning. DIL uses more face swapping suited ID features to further boost identity similarity. In the inference phase, FAL only retains $E_{att}$ for attribute extraction, making the testing process more convenient. 
   It is noted that HiFiVFS is trained and tested in the ~\textbf{latent space} ~\cite{rombach2022high}, but for visualization purposes, we illustrate all processes in the original image space.
   }
   \label{fig:ppl}
\end{figure*}

\section{Methods}
\label{sec:methods}

To generate high fidelity video face swapping results, we introduce a novel diffusion-based video face swapping framework, named  HiFiVFS. We first
introduce the overall framework in Sec.~\ref{sec:overview}, and then propose Fine-grained Attribute Learning (Sec.~\ref{sec:FAL}) introduces adversarial learning to capture detailed attribute features while mitigating the effects of identity leakage. Meanwhile, Detailed Identity Learning (Sec.~\ref{sec:EIT}) analyzes the 
difference between face recognition and face swapping tasks, 
and proposes a more effective way to preserve identity features. 

\subsection{Overview\label{sec:overview}}
As shown in Fig.~\ref{fig:ppl}, the input consists of a target video $V_{t}$ and a source image $I_{s}$. During training, $I_{s}$ is randomly selected from the frames of $V_{t}$, while during testing, $I_{s}$ can be any aligned face image. Following previous diffusion-based face swapping methods~\cite{zhao2023diffswap,liu2024towards,han2024face}, we construct the data flow around the inpainting pipeline. To further improve upon previous methods, we propose FAL to capture detailed attributes and DIL to enhance identity similarity.

 HiFiVFS is primarily built upon the SVD~\cite{blattmann2023svd} framework, which employs temporal attention and 3D convolution to ensure the stability of results. However, there are two main differences in our  HiFiVFS. First, while SVD takes a still image as input,  HiFiVFS uses multi-frame video input. This means we extend the SVD framework from an image-to-video task to a video-to-video task, making it more suitable for face swapping applications. Second, the conditioning signal in SVD is primarily a text prompt that controls the overall movement of the results. In contrast,  HiFiVFS uses an identity feature as the conditioning signal to control identity similarity, while the overall movement is influenced by the attribute information from the input video.

It is important to note that SVD uses EDM~\cite{karras2022elucidating} and predicts the expected output directly, which is different from other methods that train a separate network $F_{\theta}$ from which $D_{\theta}$ is derived. As a result, when computing losses such as identity similarity, we can directly decode the output of $D_{\theta}$ to image for our calculations, making it more advantageous for our face swapping task.


\subsection{Fine-grained Attributes Learning\label{sec:FAL}}

Previous diffusion-based methods for face swapping~\cite{zhao2023diffswap,liu2024towards, han2024face} have primarily depended on inpainting pipelines. However, these approaches often struggle to preserve subtle details, due to their limited access to detailed attribute information. To tackle this challenge, we proposed fine-grained attribute learning (FAL) to enhance the control capabilities of diffusion-based methods.

FAL includes an encoder $E_{attr}$ to extract attribute features, a decoder $Dec$ to fuse identity features with attribute features, a discriminator $Dis$ to enhance overall generation quality, and a frame mask $M_f$ to control the interaction of attribute features with Denoising-UNet. Specifically, in the FAL process, $E_{attr}$ first extracts the attribute features $f_{attr}$ from $V_t$. Then, we randomly select a face and use a pre-trained recognition model to extract its ID features $f_{rid}$. In the decoder, we fuse $f_{attr}$ and $f_{rid}$ to obtain a modified template video with a different identity $V^{'}_{t}$. Finally, we use $E_{attr}$ again to extract the attribute features $f^{'}_{attr}$ from $V^{'}_{t}$. Since the identity of $V^{'}_{t}$ has changed, $f^{'}_{attr}$ can be considered as the disentangled attribute feature. 

The FAL process requires the use of the following loss functions to ensure effectiveness. First, it is necessary to ensure that the attribute features of $V_t$ and $V^{'}_{t}$ are consistent, which can be achieved using 
\begin{equation} \label{Latt}
\mathcal{L}_{attr} = \frac{1}{2} \left\| f_{attr} - f^{'}_{attr} \right\|_2^2.
\end{equation}
To make the training process more stable, pixel-level supervision is also needed. Therefore, during the training process, $f_{rid}$ is set equal to the ID features $f_{gid}$ of $V_t$ by 50\%, allowing for the calculation of pixel-level reconstruction loss in this case, 
\begin{equation} \label{Lrec}
\mathcal{L}_{rec} = 
\begin{cases} 
\frac{1}{2} \left\| V^{'}_{t} - V_t \right\|_2^2 & \text{if } f_{rid} = f_{id}, \\
0 & \text{otherwise}.
\end{cases}
\end{equation}
To prevent the model from directly generating $V^{'}_{t}$ that is completely identical to $V_t$, we also use a triplet margin identity loss,
\begin{equation} \label{Ltid}
\mathcal{L}_{tid} = max\{\cos(f^{'}_{gid},f_{gid}) - \cos(f^{'}_{gid},f_{rid}) + m, 0 \}.
\end{equation}
Here $f_{gid}$ and $f^{'}_{gid}$  means identity feature from $V_t$ and $V^{'}_{t}$, $m$ means margin, which is set to $0.4$ during our training, $\cos$ is cosine distance. Noted that $\mathcal{L}_{tid}$ only compute when $f_{rid}$ $\neq$ $f_{gid}$. Compared with traditional identity loss, $\mathcal{L}_{tid}$ focuses more on the change of identity rather than requiring a high degree of similarity, which makes it more robust in terms of attribute preservation.

Finally, to capture fine details and further improve realism, we follow the adversarial objective in~\cite{choi2020stargan}. Thus, the overall FAL loss is formulated as:
\begin{equation} \label{eq:dal}
\mathcal{L}_{FAL} = \mathcal{L}_{adv} + \lambda_{attr}\mathcal{L}_{attr} + \lambda_{tid} \mathcal{L}_{tid} + \lambda_{rec} \mathcal{L}_{rec},
\end{equation}
with $\lambda_{attr}$ = $\lambda_{rec}$ = 10, $\lambda_{tid}$ = 1.

When injection attribute feature from FAL in SVD, we leverage the low-level features of $ E_{attr}$ and modify the injection method from the conventional cross-attention approach to directly adding these features to the input of the denoising-UNet. This adjustment aims to preserve all detailed attribute features as much as possible. 
Additionally, we implement a frame-wise mask $M_f$ to enable quick control over the attribute feature injection. During the early stages of training, $ M_f$ is fixed at zero, which effectively separates the training of FAL, ensuring robust training in the initial phase. After this warm-up stage, $M_f$ is used to randomly drop the attribute information.

It should be emphasized that the training process of FAL is similar to GAN-based face swapping methods, but there are several differences: 1) FAL places more emphasis on maintaining attributes, while only using a weaker constraint on identity to ensure it is not completely identical to the original input. 2) The structure of FAL is based on the latent space of VAE~\cite{rombach2022high}, while previous works are mainly on the pixel space. So FAL is more compatible with the training of Denoising-UNet.

\subsection{Detailed Identity Learning \label{sec:EIT}}

Previous face swapping methods primarily relied on the global 512-dimensional features $f_{gid}$ extracted from the last fully connected (FC) layer of a pre-trained face recognition model to represent identity. Face recognition focuses on distinguishing between different individuals, while $f_{gid}$ effectively captures key identity information, it often lacks finer details which is important for generation tasks.

To get more detailed identity information, we use the features from the last Res-Block layer, which has not yet been significantly compressed, called $f_{did}$. Then we propose a Detailed Identity Tokenizer (DIT) to transfer $f_{did}$ to tokens.
Specifically, we use a convolutional layer to align the number of channels in $f_{did}$ with the dimension of the cross attention in UNet. Next, we flatten and transpose this feature along the height and width dimensions, treating each spatial pixel as a token, resulting in 49 tokens $t_{did}$. These tokens are then fed into the cross attention and temporal attention modules of UNet. This approach maximizes the retention of all ID details. Additionally, to improve similarity further, we also calculated the identity loss:
\begin{equation} \label{eq:id}
\mathcal{L}_{\text{id}} = 1 - \cos(f_{gid},E_{id}(V_{r})).
\end{equation}
Noted that when calculating the ID loss, we still used the global features $f_{gid}$ to reduce the impact of extra factors like pose and expression.

\subsection{Overall Loss Function} \label{sec:loss}

Our Loss function has 4 components, including Denoising Score Matching (eq.~\ref{eq:dm}),  Fine-grained Attributes Learning Loss (eq.~\ref{eq:dal}), and Identity Loss (eq.~\ref{eq:id}). The overall Loss can be written as:
\begin{equation} \label{eq:all}
\mathcal{L} = \mathcal{L}_{\text{DM}} + \lambda_{\text{FAL}} \mathcal{L}_{\text{FAL}}  + \lambda_{\text{id}} \mathcal{L}_{\text{id}},
\end{equation}
with $\lambda_{\text{FAL}}$ = 1, $\lambda_{\text{id}}$ = 0.1.

\section{Experiments}
\label{sec:exp}
\subsection{Experimental Setup \label{sec:expsetupl}}
\noindent\textbf{Dataset.} 
We use VoxCeleb2~\cite{chung2018voxceleb2}, CelebV-Text~\cite{yu2023celebv}, and VFHQ~\cite{xie2022vfhq} as our training datasets.
We first evaluate our method using the FaceForensics++(FF++)~\cite{rossler2019faceforensics++} dataset, which comprises 1,000 videos. 
Additionally, since FF++ is generally less challenging, primarily consisting of frontal, unobstructed cases with lower resolution, we select 100 challenging videos from VFHQ~\cite{xie2022vfhq} to create a test set that is more aligned with real-world scenarios, named VFHQ-FS. These videos were excluded from the training set. VFHQ-FS contains cases with diverse poses, expressions, lighting, makeup, and occlusion.
We conducted experiments on this dataset to further demonstrate the performance of various methods in complex scenarios.

\noindent\textbf{Evaluation Metrics.}
The quantitative evaluations are conducted using the following metrics: identity retrieval accuracy (IDr.), identity similarity (IDs.), expression error (Exp.), face shape error (Shape.), gaze error (Gaze.), pose error (Pose.), video identity distance (VIDD) and Fréchet Video Distance(FVD). For IDr. and IDs., we employ a different face recognition
model CosFace~\cite{wang2018cosface} face recognition model to perform identity retrieval and calculate identity cosine similarity. To assess Exp., Shape., Gaze., and Pose., we employ a 3DMM-based face reconstruction method from Face-Adapter~\cite{han2024face} to obtain the relative coefficients and compute the Euclidean distance between the results and the targets. 
For VIDD, we follow FOS~\cite{chen2024towards} to evaluate the temporal consistency between consecutive video frames. For FVD, we follow StyleGAN-V~\cite{skorokhodov2022stylegan}, which assesses overall video quality, reflecting spatial and temporal coherence.

\noindent\textbf{Implementation Details.} During training, for each face video, we randomly extract a clip with 16 frames and align the faces using landmarks extracted by RetinaFace~\cite{deng2020retinaface}. The frames are cropped at 640 × 640, containing more background compared to traditional crop strategy, 
which ensures that the face is complete even at extreme angles. 
We employ a pre-trained face parsing model~\cite{yu2018bisenet} to predict the face area mask.
All models are trained using the AdamW optimizer, a learning rate of 1e-5, and a batch size of 8. 
The UNet is initialized by the pre-trained weights from SVD while other trainable modules are randomly initialized. We first warm-up the training for 5,000 steps with $M_f$ fixed to zero, and continue training for an additional 50,000 steps. 
To generate results for evaluation, we use 25 steps of the deterministic EDM sampler with a classifier guidance scale of 2, and the temporal co-denoising is used to weaken the detail discrepancies between different video clips.

\begin{figure*}[t]
  \centering
   \includegraphics[width=0.99\linewidth]{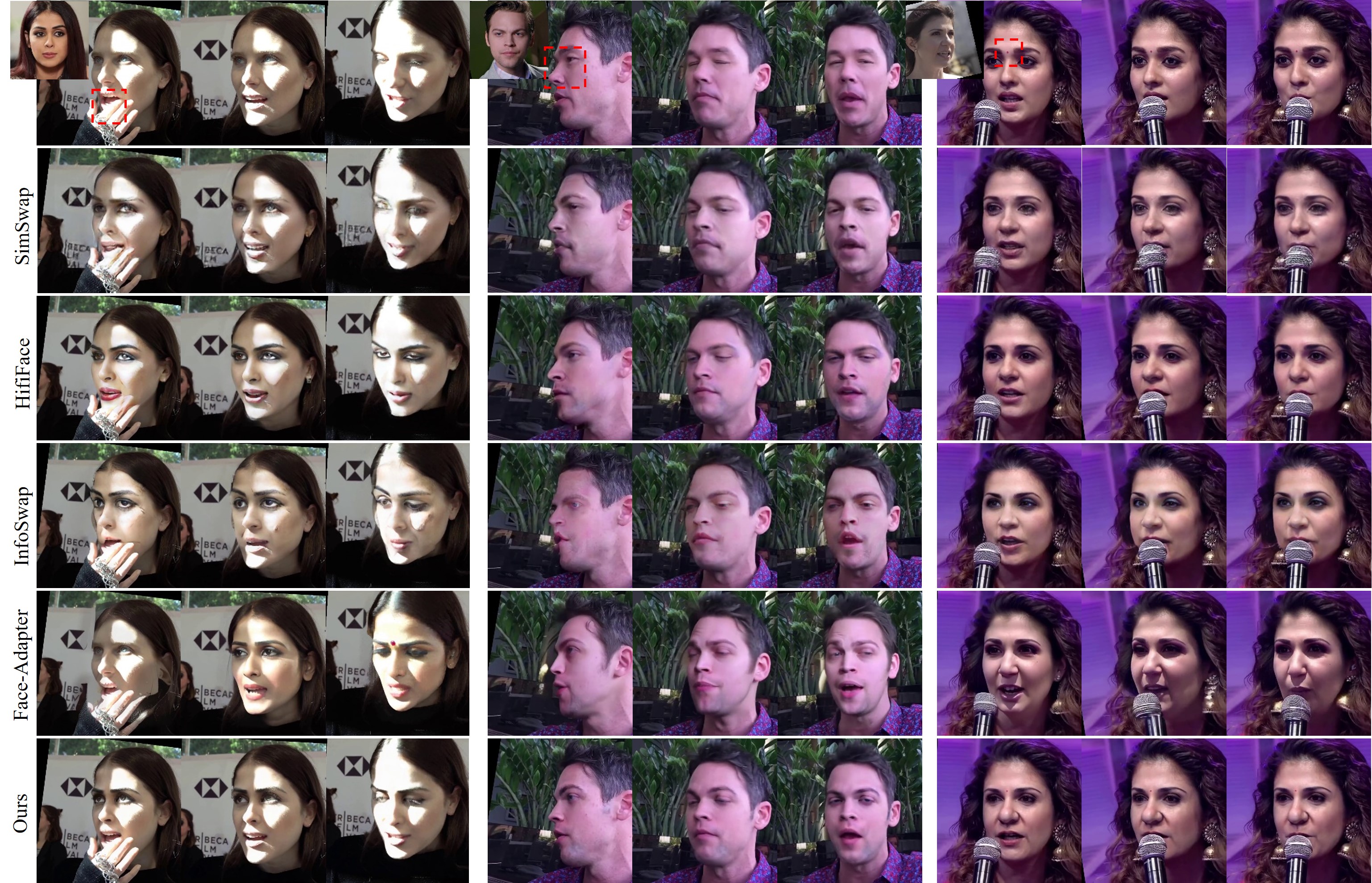}
    
   \caption{VFHQ-FS results compared with other methods. The source image of each example is placed in the corresponding top-left position, and the target videos are in the first row. The complete video comparisons are included in the \textbf{Supplementary Materials}.}
   \label{fig:vhfq-fs}
\end{figure*}

\begin{figure}[t]
  \centering
   \includegraphics[width=0.95\linewidth]{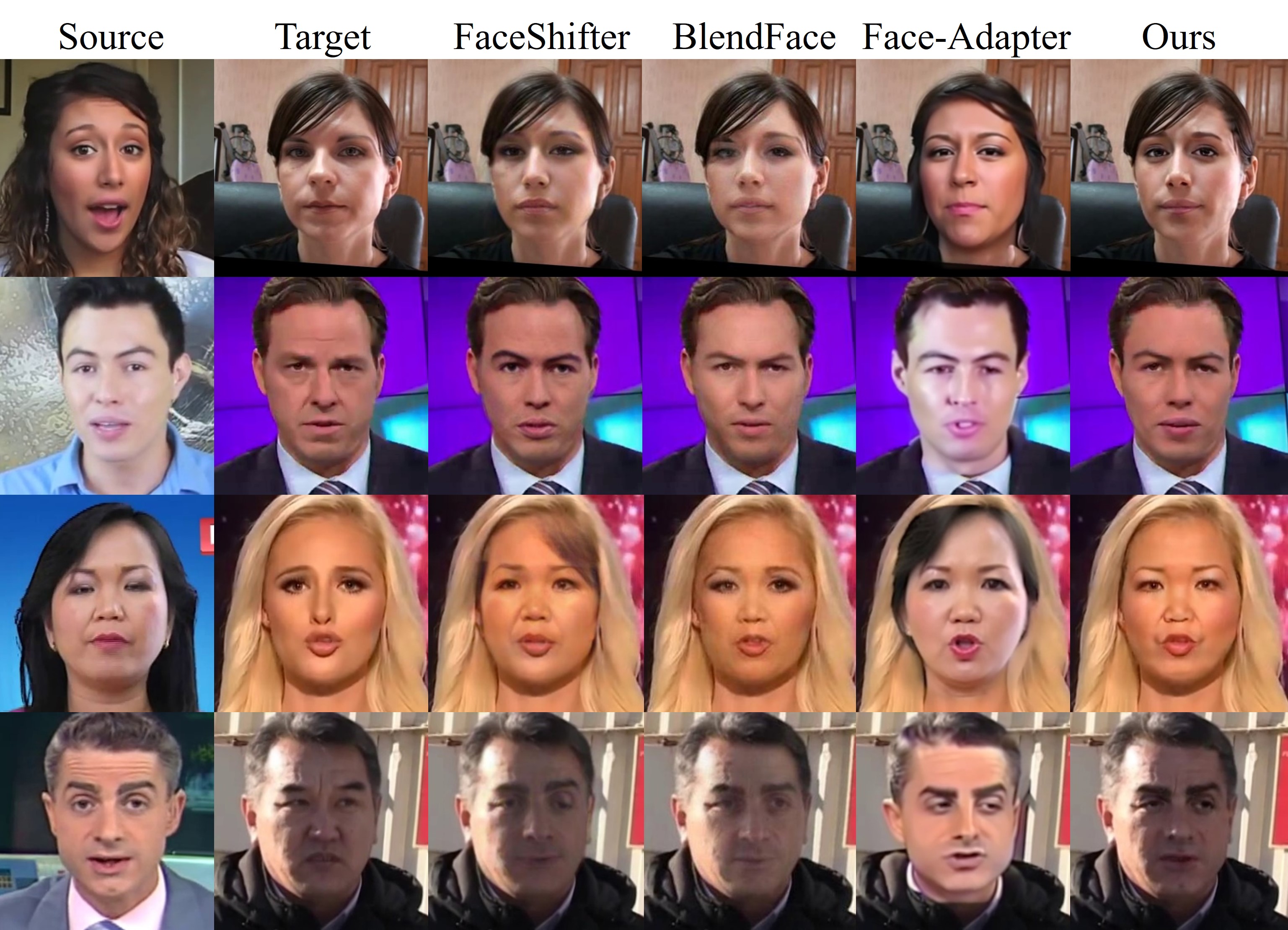} 
   \caption{FF++ results compared with FaceShifter~\cite{li2019faceshifter}, BlendFace~\cite{shiohara2023blendface} and Face-Adapter~\cite{han2024face}. }
   \label{fig:ff}
\end{figure}

\subsection{Comparisons with Existing Methods}
In this section, we compare with other methods quantitatively and qualitatively on
FF++ and VFHQ-FS test set, including GAN-based  FaceShifter~\cite{li2019faceshifter}, SimSwap~\cite{chen2020simswap}, HifiFace~\cite{wang2021hififace}, InfoSwap~\cite{gao2021information}, BlendFace~\cite{shiohara2023blendface} and diffusion-based DiffSwap~\cite{zhao2023diffswap}, Face-Adapter\cite{han2024face}.

\begin{table}[t]
\small
\caption{Quantitative results in FF++.  For DiffSwap~\cite{zhao2023diffswap}, the results from the official code differ significantly from those in the paper, so we have marked them in gray in the table.} 
\label{tab:ff}
\centering
\resizebox{0.45\textwidth}{!}{%
\begin{tabular}{ccccccc}
\toprule
Methods & IDr.$\uparrow$ & IDs.$\uparrow$   & Exp$\downarrow$ & Pose$\downarrow$ & Shape$\downarrow$ & Gaze$\downarrow$  \\
\midrule
SimSwap~\cite{chen2020simswap} &96.78 & 62.42 & 5.94& ~\textbf{0.0261}& 1.3826& 0.055  \\
HifiFace~\cite{wang2021hififace} & 94.26& 58.05 & 6.50 & 0.0382 & 1.3491 & 0.057   \\
FaceShifter~\cite{li2019faceshifter} &87.99&54.49 &6.32 &0.0342 &1.4072 &0.072 \\
InfoSwap~\cite{gao2021information} &\textbf{99.26} &67.88 &7.25 &0.0371 &1.4191 &0.062   \\
BlendFace~\cite{shiohara2023blendface} &89.91 &54.44 &6.15 &0.0286 &1.5293 &0.056   \\
\midrule
DiffSwap~\cite{zhao2023diffswap} &\textcolor{lightgray}{19.16} & \textcolor{lightgray}{28.56}&\textcolor{lightgray}{4.94} &\textcolor{lightgray}{0.0237} &\textcolor{lightgray}{2.0862} &\textcolor{lightgray}{0.067} \\
Face-Adapter~\cite{han2024face} &96.47& 53.90 &6.66 &0.0319 &1.5354 &0.061 \\
Ours &99.10 &\textbf{70.45} & \textbf{4.99}&0.0334 &\textbf{1.2243} & \textbf{0.053} \\
\bottomrule
\end{tabular}
}
\end{table}

\begin{table}[t]
\small
\caption{Quantitative results in VFHQ-FS.}
\label{tab:vfhq}
\centering
\resizebox{0.45\textwidth}{!}{%
\begin{tabular}{ccccccc}
\toprule
Methods & IDs.$\uparrow$ & Exp$\downarrow$ & Pose$\downarrow$ & Shape$\downarrow$ & VIDD$\downarrow$ &  FVD$\downarrow$ \\
\midrule
SimSwap~\cite{chen2020simswap} & 58.32 & 6.60 & 0.0499 & 1.2674 & 0.5541 &  93.59 \\ 
HifiFace~\cite{wang2021hififace} & 61.57 & 6.62 & 0.0497 &  1.2077 & 0.5599 & 101.67 \\
InfoSwap~\cite{gao2021information} & 59.80 & 7.57 & 0.0606 & 1.5067 & 0.7307 & 100.50 \\
\midrule
Face-Adapter~\cite{han2024face} & 54.82 & 6.82 & 0.0466 & 1.6172 & 0.7355 &  211.66\\
Ours & ~\textbf{63.69} & ~\textbf{5.09} & ~\textbf{0.0396} & ~\textbf{1.2066} & ~\textbf{0.5041} & ~\textbf{81.12} \\
\bottomrule
\end{tabular}
}
\end{table}

\noindent\textbf{Quantitative Comparisons.} 
We conducted experiments on the FF++ and VFHQ-FS datasets, as shown in Tables ~\ref{tab:ff} and ~\ref{tab:vfhq}. 
For the FF++, following standard practice, we sampled 10 images from each video. For the VFHQ-FS, we utilized the first 64 frames of each video.
The results indicate that HiFiVFS outperforms previous methods in most metrics, effectively generating swapped faces with improved attribute and identity control. We achieved second-place pose results on the FF++ but reached SOTA performance on the VFHQ-FS, where extreme poses are present. The small deviations in FF++ are likely due to biases from the pose estimation model. Additionally, VFHQ-FS excels in temporal stability metrics (VIDD and FVD), outperforming all methods, particularly the diffusion-based Face-Adapter~\cite{han2024face}.

\noindent\textbf{Qualitative Comparisons.} For VFHQ-FS, comparison with the previous work is shown in Fig.~\ref{fig:vhfq-fs}. 
The first example showcases the performance of our method in handling complex occlusions and lighting conditions. In contrast, the comparison methods exhibit significant artifacts, particularly in line 1, where most methods fail to accurately render the mouth occlusions. The second example focuses on extreme pose scenarios, revealing that the results from SimSwap~\cite{chen2020simswap} and InfoSwap~\cite{gao2021information} appear unnatural in this context. The third example emphasizes makeup details, such as the red dot on the forehead, which the comparison methods struggle to preserve, while our method effectively maintains this detail.
Regarding video stability, as it is challenging to assess from still images, we strongly encourage readers to refer to the \textbf{Supplementary Materials} to view the original videos. The video results clearly demonstrate that our method significantly outperforms others in terms of both video stability and generation quality. We also perform the qualitative evaluation on the FF++ dataset in Fig.~\ref{fig:ff}, and the complete comparison with more methods is shown in supplementary materials. The results of our HiFiVFS demonstrate strong identity performance while effectively preserving attributes such as expression, posture, and lighting.

\begin{figure*}[t]
  \centering
   \includegraphics[width=0.99\linewidth]{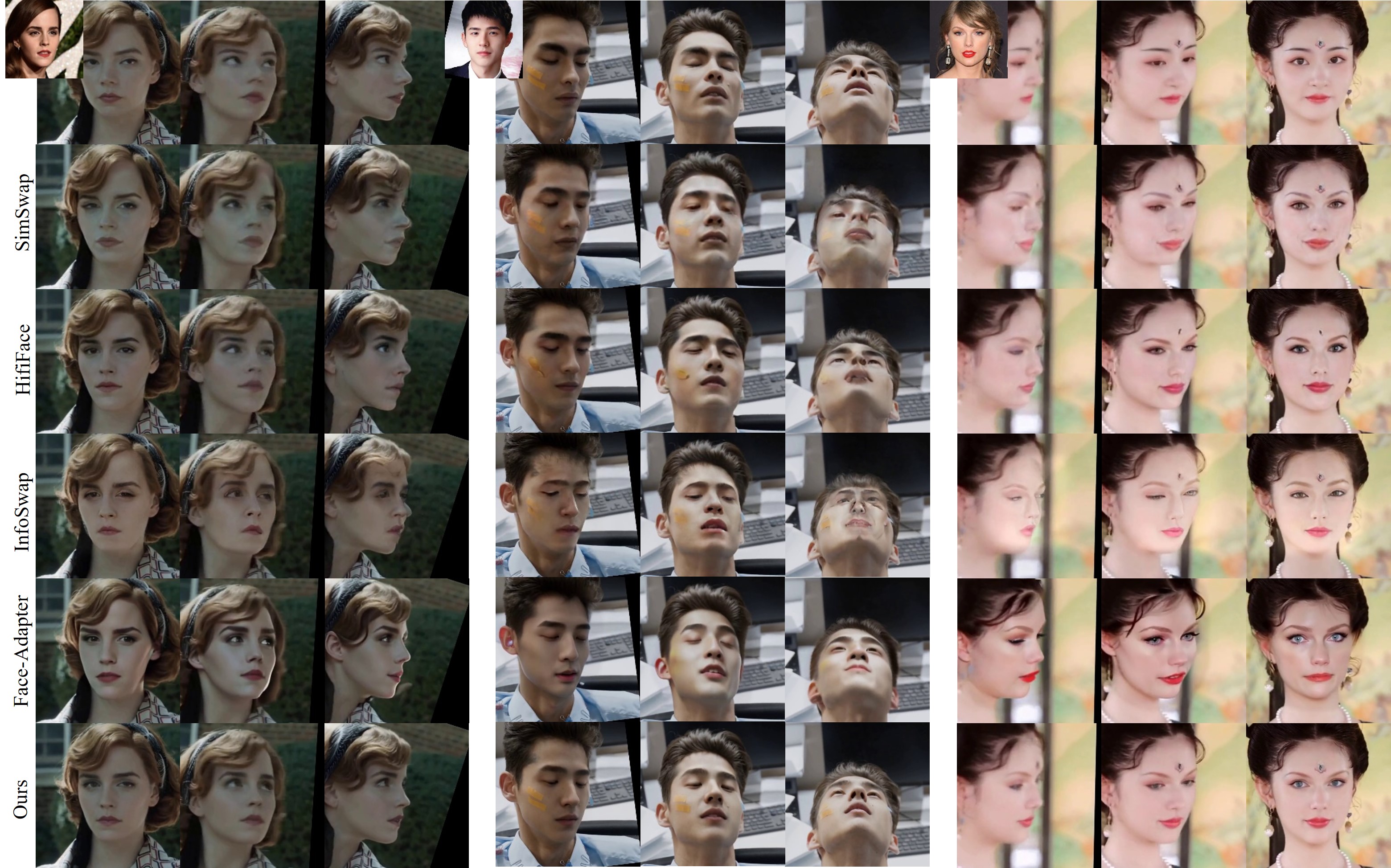}
    
   \caption{Face swapping results on wild face videos under various challenging conditions. Our method is capable of producing results with high identity similarity, excellent preservation of detailed attributes, and temporal stability.}
   \label{fig:wild}
\end{figure*}

\noindent\textbf{Human Evaluation.} 
Four user studies were conducted to evaluate the performance of the proposed model. Participants were asked to rate the results on a score from 1 (the worst) to 5 (the best): 1) identity similarity with the source face; 2) preservation of target video attributes, including pose, expression, gaze, lighting, and makeup; 3) stability of the resulting videos; and 4) generation quality of the outputs. In each unit, participants were presented with two real inputs: the source image and the target videos, along with four reshuffled face swapping results generated by SimSwap~\cite{chen2020simswap}, HifiFace~\cite{wang2021hififace}, InfoSwap~\cite{gao2021information}, Face-Adapter~\cite{han2024face}, and our HiFiVFS. Each user was presented with 20 randomly selected pairs from the VFHQ-FS and in-the-wild test sets.
Finally, we collected answers from 15 human evaluators.
The average scores for each method across the studies are shown in Tab.~\ref{tab:user}, indicating that our model significantly outperforms the other three methods. Parts of the used videos are in the supplementary materials.

\begin{table}[t]
\small
\caption{User study results of average scores for each method.}
\label{tab:user}
\centering
\resizebox{0.40\textwidth}{!}{%
\begin{tabular}{ccccc}
\toprule
Methods & ID$\uparrow$ & Attr$\uparrow$ & Stability$\uparrow$ & Quality$\uparrow$ \\
\midrule
SimSwap~\cite{chen2020simswap} &3.35&3.57 &3.38 &3.28   \\
HifiFace~\cite{wang2021hififace} &3.92&3.55 &3.52 &3.47  \\
InfoSwap~\cite{gao2021information} &3.32&3.07 &3.08 &3.03   \\
\midrule
Face-Adapter~\cite{han2024face} &3.82&2.68 &1.57 &2.17  \\
Ours & \textbf{4.27}&\textbf{4.38} &\textbf{4.58} &\textbf{4.53}  \\
\bottomrule
\end{tabular}
}
\end{table}

\subsection{Ablation Study}

\noindent\textbf{Fine-grained Attribute Learning.}
To verify the effectiveness of FAL, we compared it with three baseline models: 1) w/o-latent, which trains FAL in pixel space instead of latent space, 2) ID+, which uses traditional identity cosine similarity loss
instead of the triplet margin identity loss employed in FAL, and 3)w/o-adv, which does not train with a discriminator. As shown in Tab.~\ref{tab:dit}, 
The w/o-latent model has a decline in all metrics. This may highlight the gap between image space and latent space. Since the denoising-UNet operates in the latent space, the FAL may require significantly more training to achieve comparable results to ours when training in image space. Both ID+ and w/o-adv have demonstrated a decrease in attribute-related metrics, indicating that triplet margin identity loss and adversarial learning are beneficial for fine-grained attribute learning.

\noindent\textbf{Detailed Identity Learning.}  
To evaluate the effectiveness of DIL, we train models with $f_{gid}$ instead of $f_{did}$, which is named w/o-$f_{did}$. 
As shown in Tab.~\ref{tab:dit},  w/o-$f_{did}$ achieves similar performance in attribute metrics; however, it performs less effectively in terms of identity similarity and face shape. The results prove that the using of $f_{did}$ is beneficial to the face swapping task.

\begin{table}[t]
\small
\caption{Ablation Study of FAL and DIL on FF++. Bold text represents the best and underlined text represents the second result.}
\label{tab:dit}
\centering
\resizebox{0.40\textwidth}{!}{%
\begin{tabular}{ccccccc}
\toprule
Methods  & IDs.$\uparrow$   & Exp$\downarrow$ & Pose$\downarrow$ & Shape$\downarrow$ & Gaze$\downarrow$  \\
\midrule
w/o-latent & 69.95 & 5.50 & 0.0392 & 1.2587 & 0.062   \\
ID+ & \textbf{72.02} & 5.46 & 0.0365 & \textbf{1.2217} & 0.055     \\
w/o-adv& 70.37 & 5.22 & 0.0356 & 1.2293 & 0.056 \\
w/o-$f_{did}$& 68.37 & \underline{5.04} & \underline{0.0335} & 1.2548 & \textbf{0.051}  \\
Ours &\underline{70.45} & \textbf{4.99}&\textbf{0.0334} &\underline{1.2243} & \underline{0.053} \\
\bottomrule
\end{tabular}
}
\end{table}

\section{Conclusion}
In this work, we introduce a high fidelity video face swapping method called HiFiVFS, which can generate high fidelity face swapping videos even in extremely challenging scenarios. HiFiVFS leverages the generative and temporal priors of SVD to build the first true video face swapping framework. FAL achieves attribute disentanglement and enhancement through identity desensitization and adversarial learning, thereby improving control over fine-grained attributes. DIL extracts detailed identity features for the face swapping task, thereby improving the identity similarity. Extensive experiments demonstrate that HiFiVFS significantly outperforms previous face swapping methods.

{
    \small
    \bibliographystyle{ieeenat_fullname}
    \bibliography{main}
}
\clearpage
\setcounter{page}{1}
\maketitlesupplementary

\section{Network Structures of FAL}
\begin{figure}[t]
  \centering
   \includegraphics[width=0.8\linewidth]{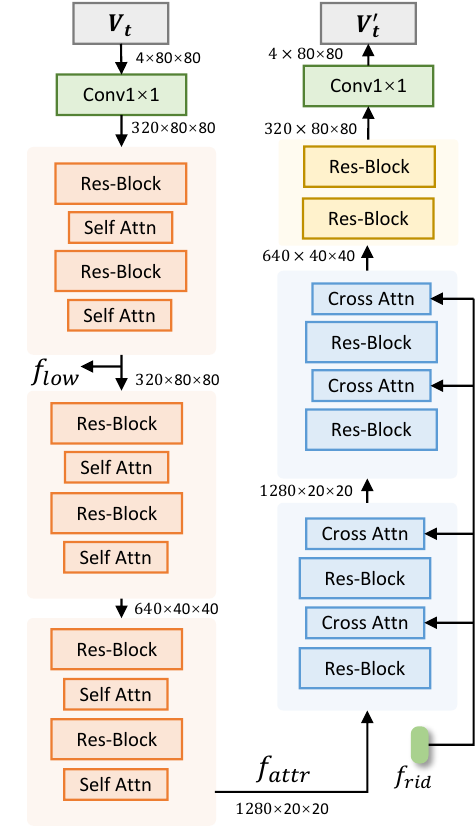}
    
   \caption{Detail Network Structures of FAL}
   \label{fig:structure}
\end{figure}

The detailed structure of our HiFiVFS is shown in Fig.~\ref{fig:structure}. The encoder $E_{attr}$ consists of three layers, each containing two residual blocks and two self-attention mechanisms. The output $f_{attr}$ is obtained from the last layer of $E_{attr}$, while $f_{low}$ represents low-level features that are combined with the denoising-UNet to preserve all relevant attribute features as much as possible. The decoder also has three layers, with the first two layers employing cross-attention to merge $f_{rid}$. The discriminator shares the same structure as $E_{attr}$ but includes an additional convolution layer at the end to adjust the output channels to 2.

\section{More Results}
\begin{figure*}
  \centering
  \begin{subfigure}{0.95\linewidth}
    \centering
    \includegraphics[width=0.8\linewidth]{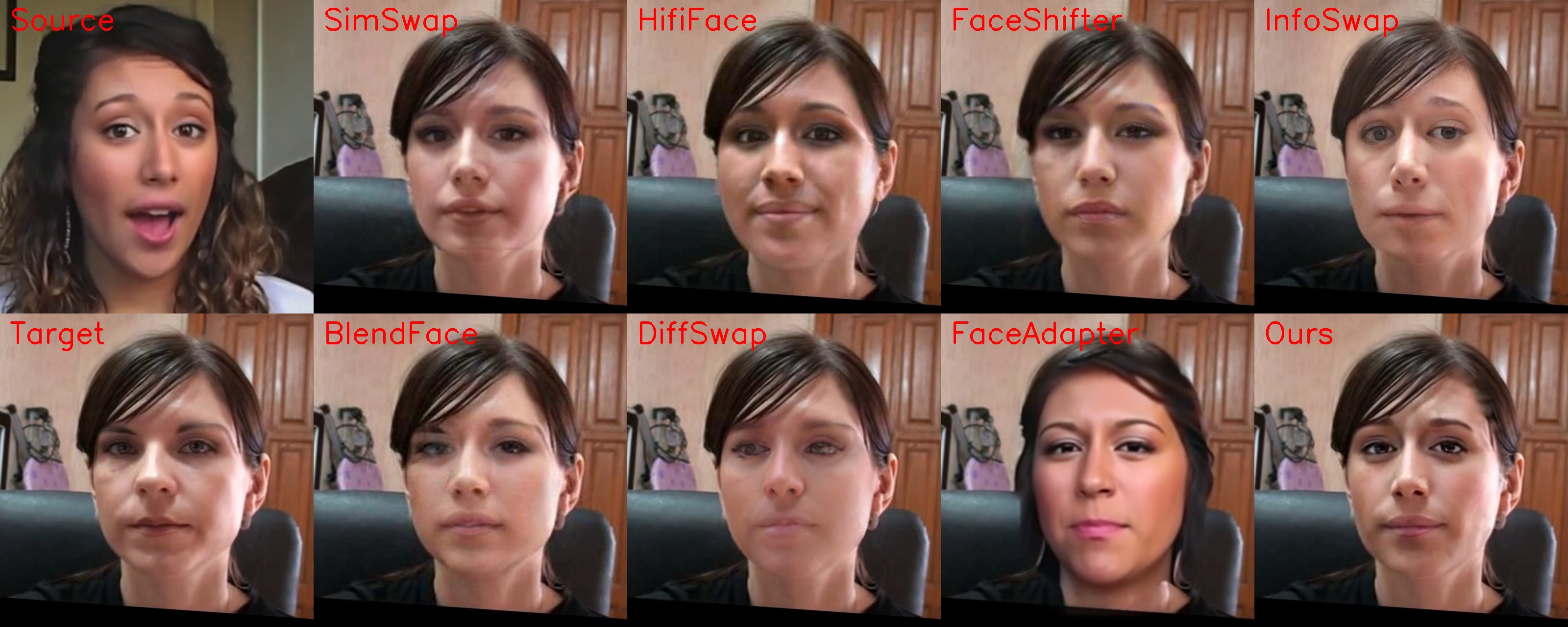}
  \end{subfigure}
  \vfill
  \begin{subfigure}{0.95\linewidth}
    \centering
    \includegraphics[width=0.8\linewidth]{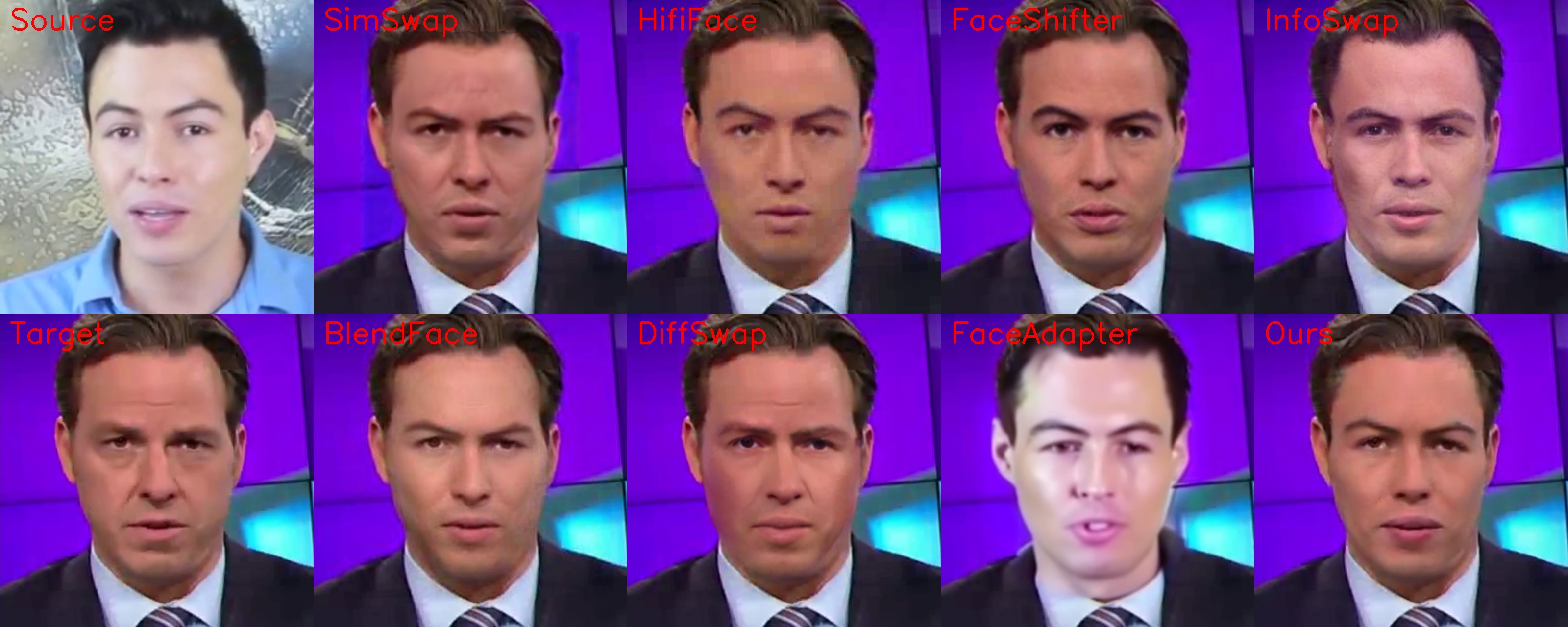}
  \end{subfigure}
  \vfill
  \begin{subfigure}{0.95\linewidth}
    \centering
    \includegraphics[width=0.8\linewidth]{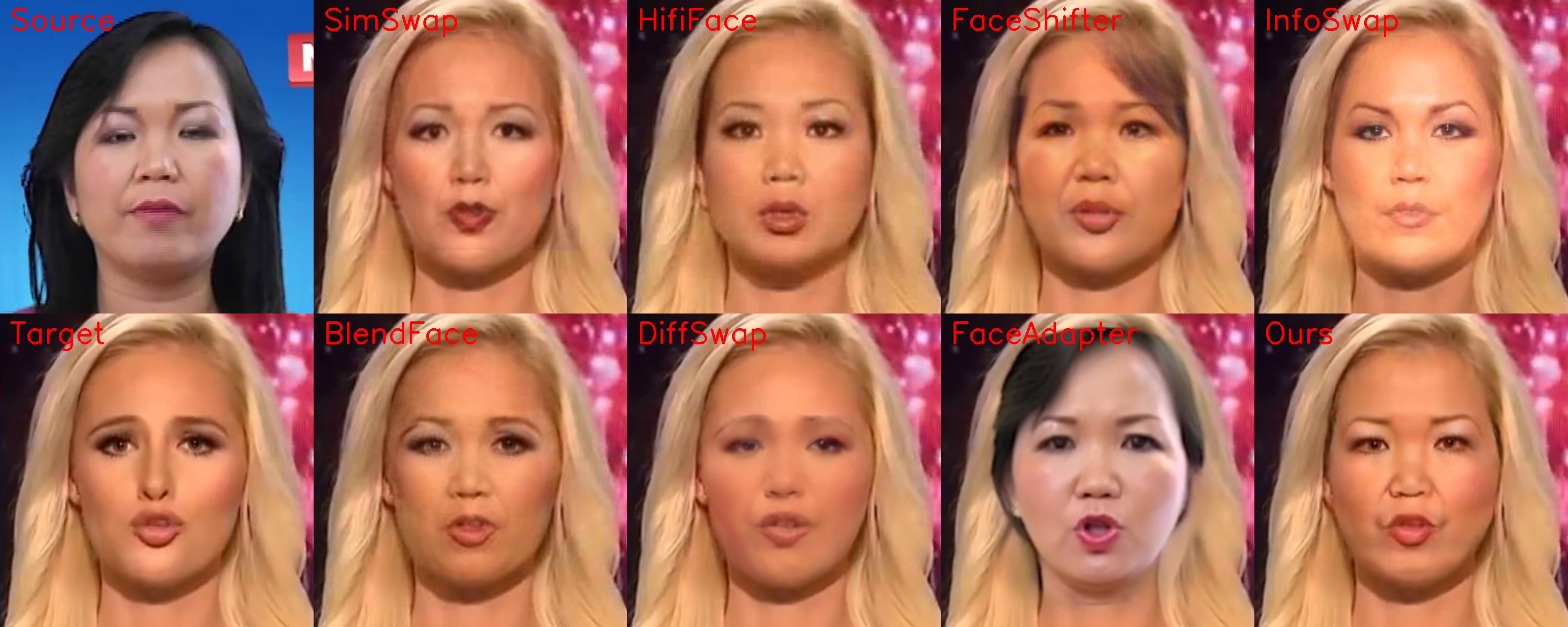}
  \end{subfigure}
  \vfill
  \begin{subfigure}{0.95\linewidth}
    \centering
    \includegraphics[width=0.8\linewidth]{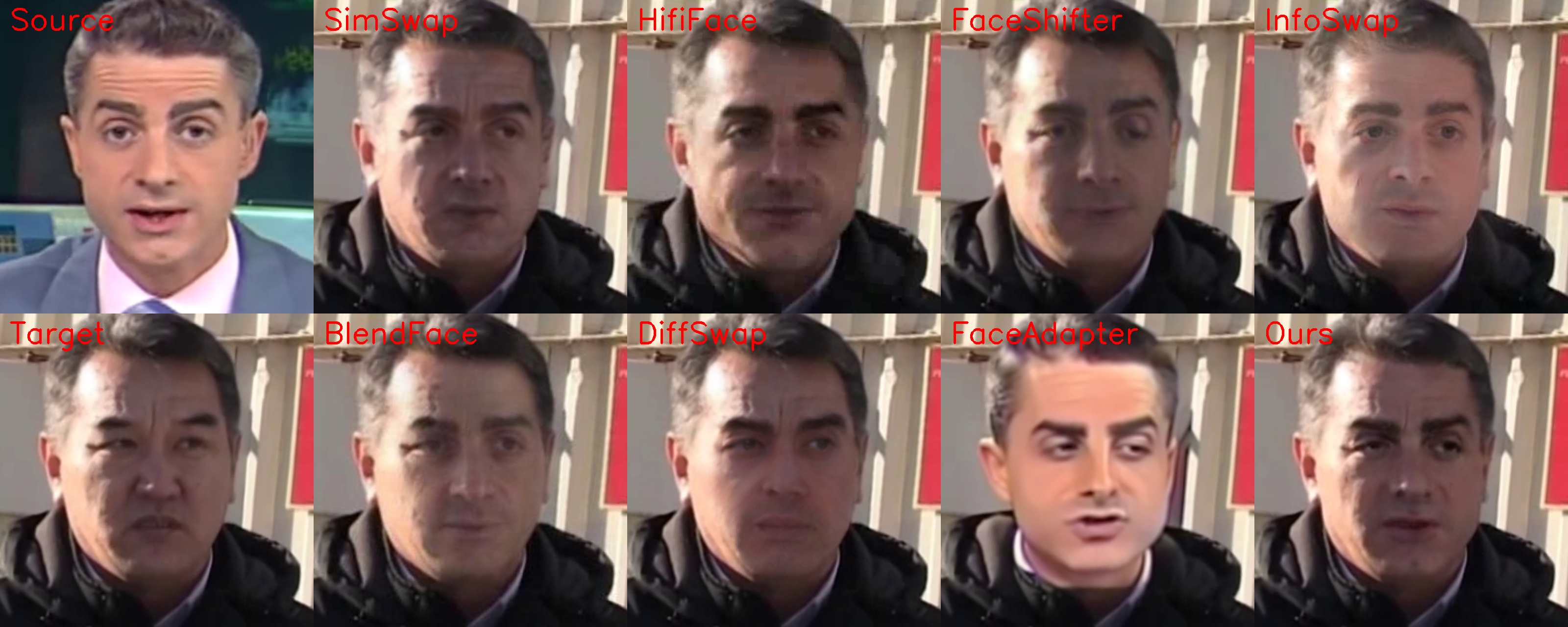}
  \end{subfigure}
  \caption{More comparisons on FF++.}
  \label{fig:sup_ff++1}
\end{figure*}

\begin{figure*}
  \centering
  \begin{subfigure}{0.95\linewidth}
    \centering
    \includegraphics[width=0.8\linewidth]{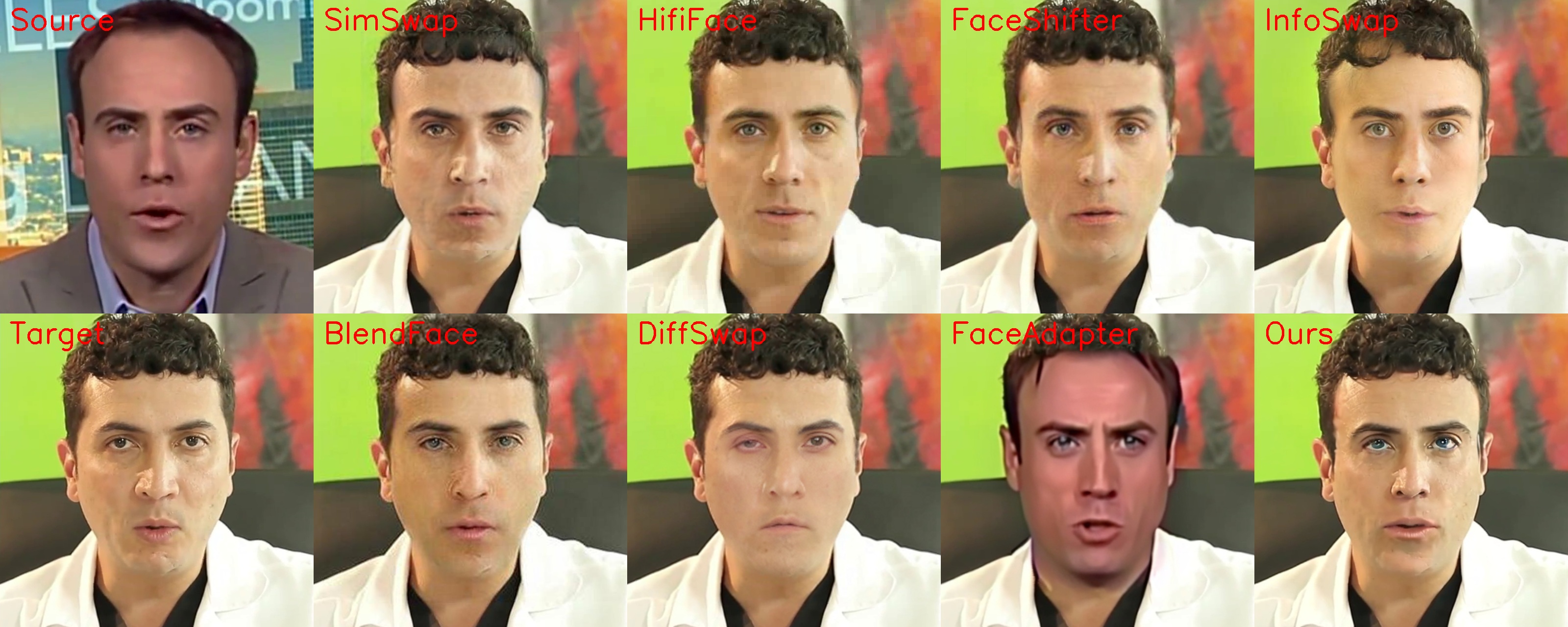}
  \end{subfigure}
  \vfill
  \begin{subfigure}{0.95\linewidth}
    \centering
    \includegraphics[width=0.8\linewidth]{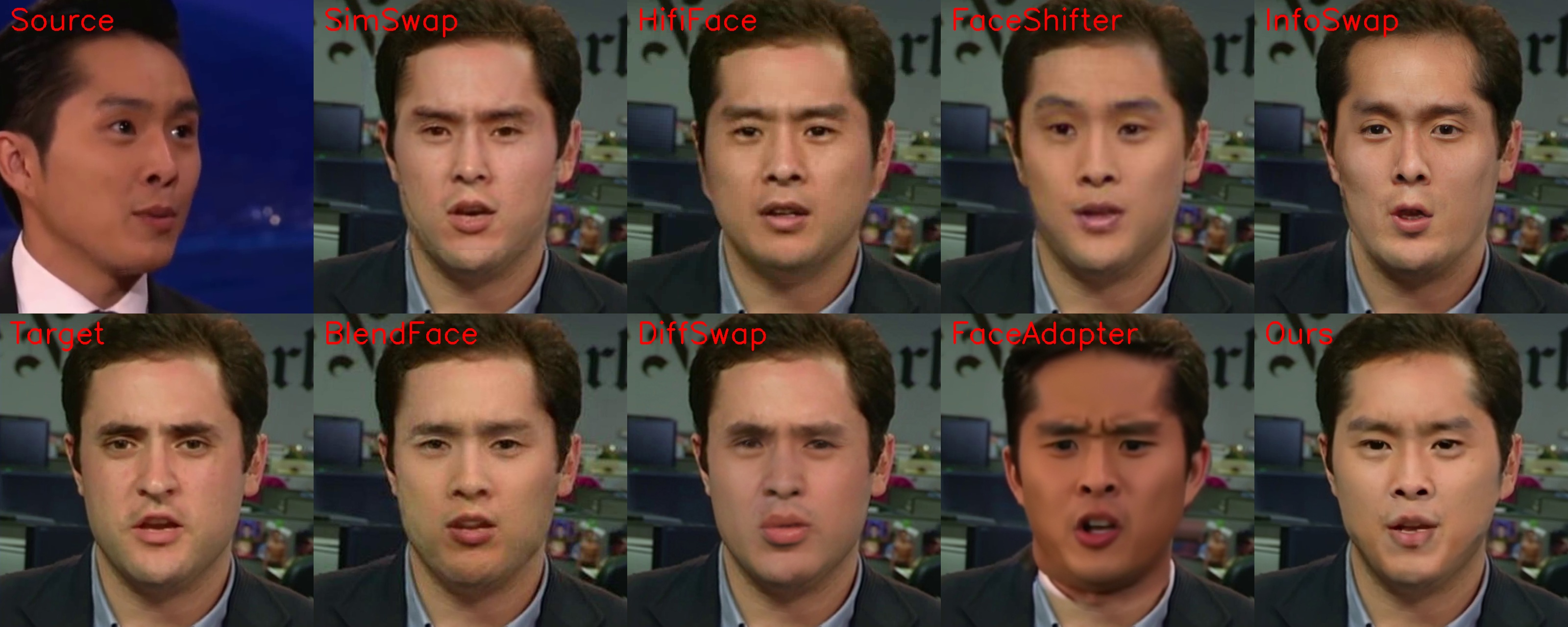}
  \end{subfigure}
  \vfill
  \begin{subfigure}{0.95\linewidth}
    \centering
    \includegraphics[width=0.8\linewidth]{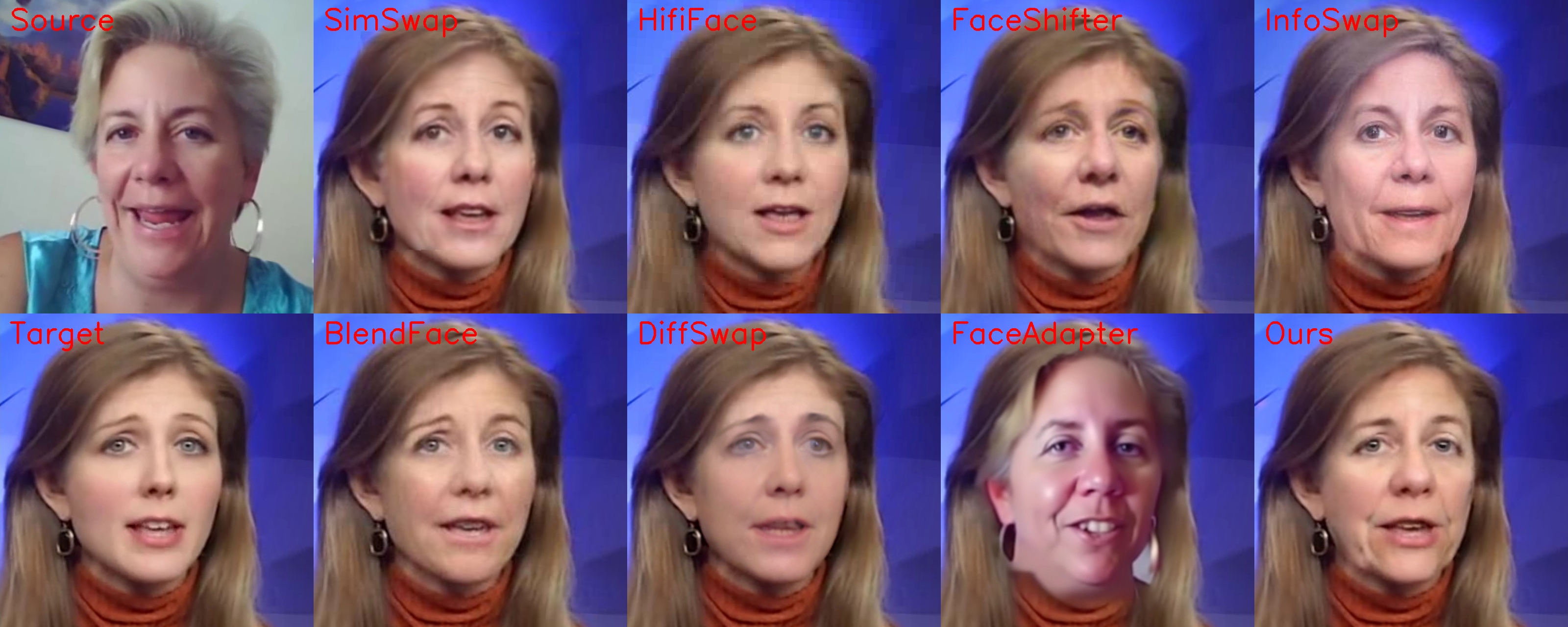}
  \end{subfigure}
  \vfill
  \begin{subfigure}{0.95\linewidth}
    \centering
    \includegraphics[width=0.8\linewidth]{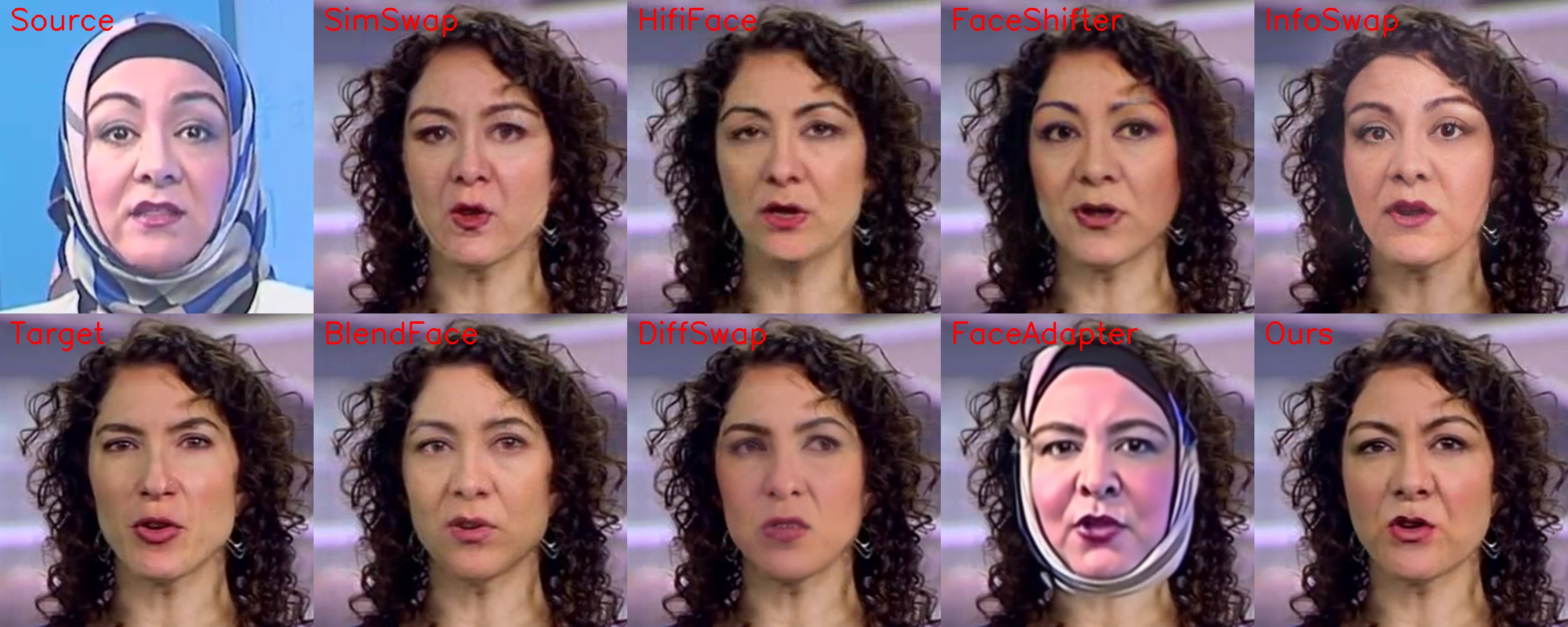}
  \end{subfigure}
  \caption{More comparisons on FF++.}
  \label{fig:sup_ff++2}
\end{figure*}

For FF++, we put more comparations in Fig.~\ref{fig:sup_ff++1} and ~\ref{fig:sup_ff++2}. For VFHQ-FS and wild cases, we have included additional comparative results in the zip file (Comparisons\_VHFQ-FS and Comparisons\_Wild). Besides the academic methods mentioned in the main text, we also conducted a comparison with the wildly-used open-source tool DeepFaceLive in wild cases. HiFiVFS demonstrates superior performance compared to other state-of-the-art face swapping methods on wild face videos across a variety of scenarios.

Additionally, we have included several videos longer than 64 frames in the Long\_Samples folder to illustrate that HiFiVFS maintains consistent performance with extended video lengths.

\section{Broader Impact and Limitations}
\label{sec:limit}

\noindent\textbf{Broader Impact.} HiFiVFS is capable of consistently producing high-quality face-swapping videos, even in highly challenging situations, which expands the potential applications of face-swapping technology. However, the risk of misuse poses serious concerns, such as privacy infringements, the dissemination of false information, and various ethical dilemmas. To mitigate these risks, it is essential to thoroughly assess the models, their intended uses, safety implications, associated risks, and potential biases before implementing them in real-world contexts. On a positive note, HiFiVFS can also play a significant role in forgery detection, enhancing our ability to recognize and combat deepfakes.

\noindent\textbf{Limitation.} Video diffusion models are generally slow in sampling and demand significant VRAM, and HiFiVFS is no exception. To address this challenge, diffusion distillation methods present a promising solution for achieving faster synthesis.

\end{document}